\documentclass[preprint,12pt,5p,times,twocolumn]{elsarticle}
\journal{Operations Research Letters}
\usepackage{amssymb, amsmath, amsthm}
\usepackage[dvipsnames]{xcolor}
\usepackage{hyperref}
\usepackage[ruled,norelsize]{algorithm2e}
\SetKwInOut{Hyperparameters}{Hyperparameters}
\usepackage{enumitem}
\usepackage{subcaption}
\usepackage{booktabs} 
\usepackage{siunitx}
\hypersetup{
	breaklinks=true,   
	colorlinks=true,   
	pdfusetitle=true,  
}
\PassOptionsToPackage{linktocpage,hyperindex,breaklinks}{hyperref}
\usepackage{import}
\usepackage{pkgs}
\usepackage[hyphenbreaks]{breakurl}
\usepackage{url}
\PassOptionsToPackage{hyphens}{url}

\theoremstyle{definition}

\begin{document}

\begin{frontmatter}
\title{Structured Pruning of Neural Networks for Constraints Learning}

\author[poly]{Matteo Cacciola}
\author[unipi]{Antonio Frangioni}
\author[cornell]{Andrea Lodi}

\affiliation[poly]{
            organization={CERC, Polytechnique Montréal},
            city={Montréal},
            country={Canada}}
\affiliation[unipi]{
            organization={University of Pisa},
            city={Pisa},
            country={Italy}}
            
\affiliation[cornell]{
            organization={Cornell Tech and Technion -- IIT},
            city={New York},
            country={USA}}

\address{matteo.cacciola@polymtl.ca\\
frangio@di.unipi.it\\
andrea.lodi@cornell.edu}

\begin{abstract}
In recent years, the integration of Machine Learning (ML) models with Operation Research (OR) tools has gained popularity across diverse applications, including cancer treatment, algorithmic configuration, and chemical process optimization. In this domain, the combination of ML and OR often relies on representing the ML model output using Mixed Integer Programming (MIP) formulations. Numerous studies in the literature have developed such formulations for many ML predictors, with a particular emphasis on Artificial Neural Networks (ANNs) due to their significant interest in many applications. However, ANNs frequently contain a large number of parameters, resulting in MIP formulations that are impractical to solve, thereby impeding scalability. In fact, the ML community has already introduced several techniques to reduce the parameter count of ANNs without compromising their performance, since the substantial size of modern ANNs presents challenges for ML applications as it significantly impacts computational efforts during training and necessitates significant memory resources for storage. In this paper, we showcase the effectiveness of pruning, one of these techniques, when applied to ANNs prior to their integration into MIPs. By pruning the ANN, we achieve significant improvements in the speed of the solution process. We discuss why pruning is more suitable in this context compared to other ML compression techniques, and we identify the most appropriate pruning strategies. To highlight the potential of this approach, we conduct experiments using feed-forward neural networks with multiple layers to construct adversarial examples. Our results demonstrate that pruning offers remarkable reductions in solution times without  hindering the quality of the final decision, enabling the resolution of previously unsolvable instances.
\end{abstract}

\end{frontmatter}

\section{Introduction}

The concept of embedding learned functions inside Mixed Integer Programming (MIP) formulations, also known as ``Learning-Symbolic Programming'' or ``Constraint Learning'', has gained attention in recent literature \cite{serra,munoz,anderson}. Furthermore, there has been an increase in the availability of tools that automatically embed commonly used predictive models into MIPs \cite{entmoot,janos,opticl,omlt}. These techniques and tools are especially valuable when employing ML models for predictions and utilizing OR methods for decision making based on those predictions. Unlike the traditional two-stage approaches \cite{ferreira}, embedding the predictive model within the decision-making process in an end-to-end optimization framework has been shown to yield superior results. Examples of applications are automatic algorithmic configuration \cite{iommazzo1,iommazzo2}, adversarial examples identification \cite{fischettiJo}, cancer treatments development \cite{bertsimas}, and chemical process optimization \cite{chemical1,chemical2}. 

A very relevant case is when the learned function is an ANN, since ANNs are the state-of-the-art models for numerous essential ML tasks in Computer Vision and Natural Language processing. Consequently, there have been efforts in the literature to automate the embedding of ANNs \cite{relumip}. For instance, \cite{janos}  enables to incorporate feed-forward architectures with ReLU activation functions into MIPs, utilizing the output of the ANN in the objective function. The maturity of the field is demonstrated by the fact that one of the leading commercial MIP solvers, Gurobi, recently released a \href{https://github.com/Gurobi/gurobi-machinelearning}{package} that allows feed-forward ReLU networks to be part of MIP formulations, with compatibility for popular ML packages such as PyTorch, Keras, and scikit-learn.

Unfortunately, even when we consider simple architectures that have only ReLU activation functions, the representation of an ANN in a MIP will introduce binary variables, due to the combinatorial nature of the ReLU function. Additionally, the number of binary variables and the associated constraints that need to be added to the MIP is proportional to the number of parameters in the ANN. Deep Learning has witnessed a clear trend towards developing architectures with a very large number of parameters, which contributes to ANNs high predictive power and state-of-the-art performance in various applications. This, however, poses issues in terms of training costs, storage requirements, and prediction time. Consequently, numerous methods, known as model compression techniques, have been developed to reduce the size of ANNs without compromising their predictive capability. Yet, the large size of the ANNs presents an even more significant scalability challenge when it is embedded into a MIP, due to the potentially exponential growth of the latter computational cost with its size (and, in particular, the number of binary variables). Using a state-of-the-art network in a MIP formulation may easily result in an overwhelming number of binary variables and constraints, rendering the models unsolvable within a reasonable time using any available solver.
 
In this paper, we demonstrate that pruning methods, originally developed to address specific ML challenges, can be effectively applied in the context of embedding ANNs into MIPs. Specifically, we utilize a structured pruning technique that we previously developed to significantly accelerate the solution time for adversarial example identification problems using Gurobi. 

The remainder of the paper is organized as follows: Section~\ref{section:embed} provides a formal definition of the problem concerning the embedding of learned functions in MIP formulations. Additionally, it presents one of the existing formulations from the literature specifically designed for embedding ANNs. In Section~\ref{sec:pruning}, we introduce pruning techniques and we describe the specific pruning method employed in our experiments. Section~\ref{sec:pruningformip} focuses on the benefits of pruning when incorporating ANNs into MIPs. We discuss the reasons why pruning is advantageous in this context and provide insights on selecting appropriate pruning techniques. Finally, in Section~\ref{sec:experiments} we present numerical results to empirically validate that pruning can effectively speed up the solution process of MIPs with embedded ANNs.
 
\section{Embedding learned functions in Mixed Integer Programs}\label{section:embed}

We consider a general class of (Mixed-Integer) Nonlinear Programs with ``learned constraints''. That is, the formulation of the problem would need to involve some functions $g_i(x)$, $i = 1, \ldots, k$, defined on the variable space of the optimization decisions, that are ``hard'' in the sense that no compact algebraic formulation, and not even an efficient computation oracle, is available. Yet, (large) data sets are available, or can be constructed, of outputs $\bar{y} = g_i(\bar{x})$ for given $\bar{x}$. These datasets can be used in several existing ML paradigms (Support Vector Machines, Decision Trees, ANNs, \ldots) to construct estimates $\bar{g}_i(x)$ of each $g_i(x)$, $i = 1, \ldots, k$, with a workable algebraic description that can then be inserted into an optimization model. Thus, we consider the class of Mathematical Programs with Learned Constraints (MPLC)
\begin{align}
 \min\; & cx + by & \label{eq:genof} \\
 \text{s.t.}\;
 & y_i = \bar{g}_i(x) & i = 1, \ldots, k \label{eq:gi} \\
 & A x + B y \leq d & \label{eq:other} \\
 & x \in X & \label{eq:int}
\end{align}
Linearity in \eqref{eq:genof} and \eqref{eq:other} is not strictly necessary in our development, but it is often satisfied in applications (see, e.g., \cite{janos, fischettiJo, bertsimas}) and we assume it for notational simplicity. Indeed, when $X$ in \eqref{eq:int} contains integrality restrictions on (some of) the $x$ variables, the class already contains Mixed-Integer Linear Programs (MILP), whose huge expressive power does not need to be discussed.
Of course, a significant factor in the complexity of \eqref{eq:genof}--\eqref{eq:int} is the algebraic form of the $\bar{g}_i(x)$, which impacts the class of optimization problems it ultimately belongs to. A significant amount of research is already available on formulations for embedding feedforward ANNs, in particular with ReLU activations, in a MIP context \cite{fischettiJo, serra, munoz, anderson, huchette}. In these formulations, the neural network is constructed layer by layer. Denoting the input vector at layer $\ell$ as $o_\ell$, and the corresponding weight matrix and bias vector as $W_\ell$ and $b_\ell$, respectively, one has
\[
 o_{\ell+1} = \max(\,0\,,\,W_\ell o_\ell + b_\ell\,)
\]
that can be expressed in a MI(L)P form as
%
%
%
\begin{align}
 & v^+_\ell - v^-_\ell = W_\ell o_\ell + b_\ell \label{scalprod}\\
 & 0 \leq v^+_\ell \leq M^+ z_\ell \label{zdef}\\
 & 0 \leq v^-_\ell \leq M^- ( 1 - z_\ell ) \label{zdef2}\\
 %
 %
 %
 & o_{\ell+1} = v^+_\ell \label{outdef}\\
 & z_\ell \in \{ \, 0 \,,\, 1 \, \}^m \label{bin}
\end{align}
Constraints \eqref{zdef} and \eqref{zdef2} ensure that both $v^+_\ell$ and $v^-_\ell$ are (component-wise) positive, and since the $z_\ell$ are (component-wise) binary, that at most one of them is positive. Consequently, constraint \eqref{scalprod} forces the relations $v_\ell^+ = \max\{ \, W_\ell o_\ell + b_\ell \,,\, 0 \, \}$ and $v_\ell^- = \min\{ \, W_\ell o_\ell + b_\ell \,,\, 0 \, \}$ (of course, constraint \eqref{outdef} is only there to make apparent what the output of the layer is). Denoting by $n$ the number of neurons in layer $\ell$ and by $m$ the number of neurons in layer $\ell + 1$, system \eqref{scalprod}--\eqref{bin} contains $m$ binary variables, $n+2m$ continuous variables, and $3m$ constraints. A significant aspect of this model (fragment) is the use of big-M constraints \eqref{zdef} and \eqref{zdef2}. It is well known that the choice of the value for the constants $M$ can significantly impact the time required to solve an instance. Indeed, the Optimized Big-M Bounds Tightening (OBBT) method has been developed in \cite{fischettiJo} to find effective values for this constant.

As previously mentioned,
%
%
the state-of-the-art solver Gurobi now includes an open-source Python package that automatically embeds ANNs with ReLU activation into a Gurobi model. Additionally, starting from the 10.0.1 release, Gurobi has the capability to detect if a model contains a block of constraints representing the relationship $y = g(x)$, where $g(\cdot)$ is an ANN, in order to then apply the aforementioned OBBT techniques to enhance the solution process. Despite showing a substantial improvement with respect to the previous version, the capabilities of Gurobi to solve these MIPs are still limited. In particular, when embedding an ANN into a MIP, Gurobi is not able to solve the problem in a reasonable time unless the number of layers and neurons in the ANN is small.

\section{Artificial Neural Networks Pruning}
\label{sec:pruning}
As mentioned in the introduction, the size of state-of-the-art ANNs has been growing exponentially over the years. While these models deliver remarkable performance, they come with high computational costs for training and inference, as well as substantial memory requirements for storage. To address this issue, various techniques have been developed to reduce these costs without significantly compromising the predictive power of the network. One such technique is pruning, which involves reducing the ANN size by eliminating unnecessary parameters. Consider for instance a linear layer with input $x_{inp}$, output $x_{out}$, and weight and bias tensors $W$ and $b$, i.e., $x_{out} = W x_{inp} + b$. Thus, pruning entails removing certain entries from $W$ or $b$. That is, pruning, say, the parameter $W_{1,1}$ results in the first coordinate of $x_{inp}$ being ignored in the scalar product when computing the first coordinate of $x_{out}$.

\begin{figure}
    \centering

\scalebox{0.6}{
\begin{tikzpicture}[x=2.7cm,y=1.6cm,tcancel/.append style={draw=#1, cross out, inner sep=1pt}]
  \message{^^JNeural network activation}
  \def\NI{4} 
  \def\NO{3} 
  \def\yshift{0.4} 
  
  \foreach \i [evaluate={\c=int(\i==\NI); \y=\NI/2-\i-\c*\yshift; \index=(\i<\NI?int(\i):"n");}]
              in {1,...,\NI}{ 
    \node[node in,outer sep=0.6] (NI-\i) at (0,\y) {$x_{\text{inp}}^{\index}$};
  }
  
  \foreach \i [evaluate={\c=int(\i==\NO); \y=\NO/2-\i-\c*\yshift; \index=(\i<\NO?int(\i):"m");}]
    in {\NO,...,1}{ 
    \ifnum\i=1 
      \node[node hidden]
        (NO-\i) at (1,\y) {$x_{\text{out}}^{\index}$};
      \foreach \j [evaluate={\index=(\j<\NI?int(\j):"n");}] in {1,...,\NI}{ 
        \draw[connect,white,line width=1.2] (NI-\j) -- (NO-\i);
        \draw[connect] (NI-\j) -- (NO-\i)
          node[pos=0.50] {\contour{white}{$w_{1,\index}$}};
      }
    \else 
      \node[node,blue!20!black!80,draw=myblue!20,fill=myblue!5]
        (NO-\i) at (1,\y) {$x_{\text{out}}^{\index}$};
      \foreach \j in {1,...,\NI}{ 
        \draw[connect,myblue!20] (NI-\j) -- (NO-\i);
      }
    \fi
  }
  
  \path (NI-\NI) --++ (0,1+\yshift) node[midway,scale=1.2] {$\vdots$};
  \path (NO-\NO) --++ (0,1+\yshift) node[midway,scale=1.2] {$\vdots$};


     \node[red, scale= 6.9,tcancel=red] at (NO-1) {};

  \def\agr#1{{\color{mydarkgreen}a_{#1}^{(0)}}}
  \node[below=1.65,right=0.25,mydarkblue,scale=0.9] at (NO-1)
    {$\begin{aligned} 
       &= \color{red} 0
     \end{aligned}$};
  \node[right,scale=0.9] at (1.3,-1.3)
    {$\begin{aligned}
      {\color{mydarkblue}
      \begin{pmatrix}
        \color{red}{0} \\[0.3em]
        x_{\text{out}}^{2} \\
        \vdots \\
        x_{\text{out}}^{m}
      \end{pmatrix}}
      &=
      \color{mydarkred}\sigma\left[ \color{black}
      \begin{pmatrix}
        \color{red} 0 &\color{red} 0 &\color{red} \ldots &\color{red} 0 \\
         w_{2,1} & w_{2,2} & \ldots & w_{2,n} \\
        \vdots  & \vdots  & \ddots & \vdots  \\
        w_{m,1} & w_{m,2} & \ldots & w_{m,n}
      \end{pmatrix}
      {\color{mydarkgreen}
      \begin{pmatrix}
        x_{\text{inp}}^{1} \\[0.3em]
        x_{\text{inp}}^{2} \\
        \vdots \\
        x_{\text{inp}}^{n}
      \end{pmatrix}}
      +
      \begin{pmatrix}
        \color{red}{0} \\[0.3em]
        b_{2} \\
        \vdots \\
        b_{m}
      \end{pmatrix}
      \color{mydarkred}\right]\\[0.5em]
    \end{aligned}$};
  
\end{tikzpicture}
}
    \caption{Structured pruning}
    \label{fig:struct_pruning}
\end{figure}
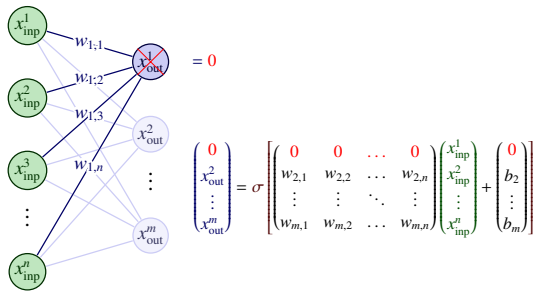

Pruning individual weight entries can offer some advantages, but it is generally suboptimal. Since most of the computation is performed on GPUs, there is little computational benefit unless entire blocks of computation, such as tensor multiplications, are removed. Removing entire structures of the ANN is known as \emph{structured} pruning, in contrast to \emph{unstructured} pruning that involves eliminating single weights. In the example of the linear layer, structured pruning would aim to remove entire neurons by deleting rows from the $W$ tensor (along with the corresponding $b$ entry in most cases). Figures~\ref{fig:unpruned_net},~\ref{fig:unstruct_pruning}, and~\ref{fig:struct_pruning}  illustrate the difference between these two pruning techniques.

\begin{figure}
\centering
\scalebox{0.6}{
\begin{tikzpicture}[x=2.7cm,y=1.6cm]
  \message{^^JNeural network activation}
  \def\NI{4} 
  \def\NO{3} 
  \def\yshift{0.4} 
  
  \foreach \i [evaluate={\c=int(\i==\NI); \y=\NI/2-\i-\c*\yshift; \index=(\i<\NI?int(\i):"n");}]
              in {1,...,\NI}{ 
    \node[node in,outer sep=0.6] (NI-\i) at (0,\y) {$x_{\text{inp}}^{\index}$};
  }
  
  \foreach \i [evaluate={\c=int(\i==\NO); \y=\NO/2-\i-\c*\yshift; \index=(\i<\NO?int(\i):"m");}]
    in {\NO,...,1}{ 
    \ifnum\i=1 
      \node[node hidden]
        (NO-\i) at (1,\y) {$x_{\text{out}}^{\index}$};
      \foreach \j [evaluate={\index=(\j<\NI?int(\j):"n");}] in {1,...,\NI}{ 
        \draw[connect,white,line width=1.2] (NI-\j) -- (NO-\i);
        \draw[connect] (NI-\j) -- (NO-\i)
          node[pos=0.50] {\contour{white}{$w_{1,\index}$}};
      }
    \else 
      \node[node,blue!20!black!80,draw=myblue!20,fill=myblue!5]
        (NO-\i) at (1,\y) {$x_{\text{out}}^{\index}$};
      \foreach \j in {1,...,\NI}{ 
        \draw[connect,myblue!20] (NI-\j) -- (NO-\i);
      }
    \fi
  }
  
  \path (NI-\NI) --++ (0,1+\yshift) node[midway,scale=1.2] {$\vdots$};
  \path (NO-\NO) --++ (0,1+\yshift) node[midway,scale=1.2] {$\vdots$};

  \def\agr#1{{\color{mydarkgreen}a_{#1}^{(0)}}}
  \node[below=1.65,right=0.25,mydarkblue,scale=0.9] at (NO-1)
    {$\begin{aligned} 
       &= \color{mydarkred}\sigma\left( \color{black}
            w_{1,1}x_{\text{inp}}^{1} + w_{1,2}x_{\text{inp}}^{2} + \ldots + w_{1,n}x_{\text{inp}}^{n} + b_1
          \color{mydarkred}\right)
     \end{aligned}$};
  \node[right,scale=0.9] at (1.3,-1.3)
    {$\begin{aligned}
      {\color{mydarkblue}
      \begin{pmatrix}
        x_{\text{out}}^{1} \\[0.3em]
        x_{\text{out}}^{2} \\
        \vdots \\
        x_{\text{out}}^{m}
      \end{pmatrix}}
      &=
      \color{mydarkred}\sigma\left[ \color{black}
      \begin{pmatrix}
        w_{1,1} & w_{1,2} & \ldots & w_{1,n} \\
        w_{2,1} & w_{2,2} & \ldots & w_{2,n} \\
        \vdots  & \vdots  & \ddots & \vdots  \\
        w_{m,1} & w_{m,2} & \ldots & w_{m,n}
      \end{pmatrix}
      {\color{mydarkgreen}
      \begin{pmatrix}
        x_{\text{inp}}^{1} \\[0.3em]
        x_{\text{inp}}^{2} \\
        \vdots \\
        x_{\text{inp}}^{n}
      \end{pmatrix}}
      +
      \begin{pmatrix}
        b_{1} \\[0.3em]
        b_{2} \\
        \vdots \\
        b_{m}
      \end{pmatrix}
      \color{mydarkred}\right]\\[0.5em]
    \end{aligned}$};
  
\end{tikzpicture}

}
    \caption{Unpruned network}
    \label{fig:unpruned_net}
\end{figure}
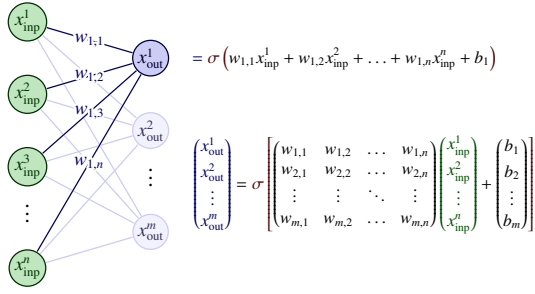

\begin{figure}
    \centering


\scalebox{0.6}{
\begin{tikzpicture}[x=2.7cm,y=1.6cm, tcancel/.append style={draw=#1, cross out, inner sep=1pt}]
  \message{^^JNeural network activation}
  \def\NI{4} 
  \def\NO{3} 
  \def\yshift{0.4} 
  
  \foreach \i [evaluate={\c=int(\i==\NI); \y=\NI/2-\i-\c*\yshift; \index=(\i<\NI?int(\i):"n");}]
              in {1,...,\NI}{ 
    \node[node in,outer sep=0.6] (NI-\i) at (0,\y) {$x_{\text{inp}}^{\index}$};
  }
  
  \foreach \i [evaluate={\c=int(\i==\NO); \y=\NO/2-\i-\c*\yshift; \index=(\i<\NO?int(\i):"m");}]
    in {\NO,...,1}{ 
    \ifnum\i=1 
      \node[node hidden]
        (NO-\i) at (1,\y) {$x_{\text{out}}^{\index}$};
      \foreach \j [evaluate={\index=(\j<\NI?int(\j):"n");}] in {1,...,\NI}{ 

      \ifnum\j=1
        \draw[connect,white,line width=1.2] (NI-\j) -- (NO-\i);
        \draw[connect] (NI-\j) -- (NO-\i)
          node[pos=0.50,tcancel=red, draw] {\contour{white}{$w_{1,\index}$}};
    \else
          \draw[connect,white,line width=1.2] (NI-\j) -- (NO-\i);
        \draw[connect] (NI-\j) -- (NO-\i)
          node[pos=0.50] {\contour{white}{$w_{1,\index}$}};
    \fi
      }
    \else 
      \node[node,blue!20!black!80,draw=myblue!20,fill=myblue!5]
        (NO-\i) at (1,\y) {$x_{\text{out}}^{\index}$};
      \foreach \j in {1,...,\NI}{ 
        \draw[connect,myblue!20] (NI-\j) -- (NO-\i);
      }
    \fi
  }
  
  \path (NI-\NI) --++ (0,1+\yshift) node[midway,scale=1.2] {$\vdots$};
  \path (NO-\NO) --++ (0,1+\yshift) node[midway,scale=1.2] {$\vdots$};

  \fill[red, opacity=0.25] ($ (NO-1) + (0.65,0.1) $) rectangle ($ (NO-1) + (0.97,-0.12) $);

  \def\agr#1{{\color{mydarkgreen}a_{#1}^{(0)}}}
  \node[below=1.65,right=0.25,mydarkblue,scale=0.9] at (NO-1)
    {$\begin{aligned} 
       &= \color{mydarkred}\sigma\left( \color{black}
            \qquad\quad w_{1,2}x_{\text{inp}}^{2} + \ldots + w_{1,n}x_{\text{inp}}^{n} + b_1^{(0)}
          \color{mydarkred}\right)
     \end{aligned}$};
  \node[right,scale=0.9] at (1.3,-1.3)
    {$\begin{aligned}
      {\color{mydarkblue}
      \begin{pmatrix}
        x_{\text{out}}^{1} \\[0.3em]
        x_{\text{out}}^{2} \\
        \vdots \\
        x_{\text{out}}^{m}
      \end{pmatrix}}
      &=
      \color{mydarkred}\sigma\left[ \color{black}
      \begin{pmatrix}
        \color{red}{0} & w_{1,2} & \ldots & w_{1,n} \\
        w_{2,1} & w_{2,2} & \ldots & w_{2,n} \\
        \vdots  & \vdots  & \ddots & \vdots  \\
        w_{m,1} & w_{m,2} & \ldots & w_{m,n}
      \end{pmatrix}
      {\color{mydarkgreen}
      \begin{pmatrix}
        x_{\text{inp}}^{1} \\[0.3em]
        x_{\text{inp}}^{2} \\
        \vdots \\
        x_{\text{inp}}^{n}
      \end{pmatrix}}
      +
      \begin{pmatrix}
        b_{1} \\[0.3em]
        b_{2}\\
        \vdots \\
        b_{m}
      \end{pmatrix}
      \color{mydarkred}\right]\\[0.5em]
    \end{aligned}$};
  
\end{tikzpicture}
}

    \caption{Unstructured pruning}
    \label{fig:unstruct_pruning}
\end{figure}
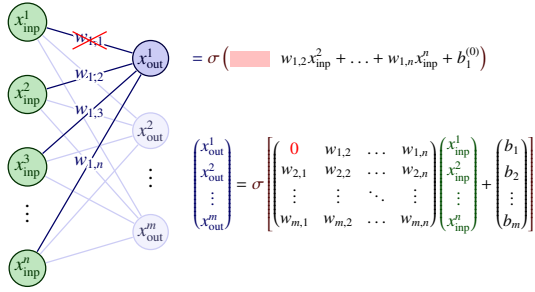

The literature on pruning techniques for neural networks is vast and encompasses a wide range of approaches. One simple and commonly used method is magnitude-based pruning, which involves removing parameters with small magnitudes. This was first introduced in \cite{han} and has been widely adopted since. However, more sophisticated strategies have also been proposed, such as Bayesian methods \cite{vdsd,drts,bcdl,acrb}, combinations of pruning with other compression techniques \cite{catd,uafc,obcf}, and zero accuracy drop pruning \cite{sipp,desp,mpcn,cbsp}.

A relevant subset of pruning techniques uses a regularization term to enforce sparsity in tensor weight. It is common practice in Machine Learning to add a regularization term $R(w)$ to the standard loss function $L(X,Y,w)$, where $w$ is the vector containing the ANNs parameters and $(X,Y)$ is the training set. Usually, $R(w)$ penalizes the magnitude of the parameters (e.g., $R(w)= |w|_2^2$) and it is known to improve the generalization performances of the model. If the form of $R(w)$ is chosen carefully, e.g., $R(w)=|w|_1$, it can also lead to a sparse parameter vector $w$. When a network parameter is zero, it can typically be removed without changing the model output for any given input. Hence, if $R(w)$ is chosen appropriately to induce all the weights of some neurons to be zero, then such neurons can be removed from the network. Many regularization terms have been proposed both for structured and unstructured pruning, including but not limited to $l_1$ norm, BerHu term \cite{berhu}, group lasso, and $l_p/l_q$ norms \cite{oto}.
%

\subsection{The Structured Perspective Regularization Term}

In the literature, the majority of pruning techniques rely on heuristics to determine the impact of removing a parameter or a structure from the ANN. This trend persists in recent works \cite{spls,tanp,wlat,spdy}, including methods that still utilize simple magnitude-based criteria \cite{dhp,temc,ampb,campfire}. Only a few techniques attempt to develop a theoretically-grounded methodology \cite{addm,lcan,lsnn,vdsd}, and these methods do not primarily focus on structured pruning.
In light of this, a pruning technique was developed in \cite{spr} that is motivated by strong theoretical foundations and specifically addresses structured pruning. In \cite{spr}, the pruning problem is addressed by starting with a na\"{\i}ve exact MIP formulation and then deriving a stronger formulation by leveraging the Perspective Reformulation technique \cite{Persp}. Analogously to what is done in \cite{FGGP11,FrFG16,FrFG17} for individual variables rather than groups of them, an efficient way to solve the continuous relaxation of this problem is obtained by projecting away the binary variables, resulting in an equivalent problem to standard ANN training with the inclusion of the new \emph{Structured Perspective Regularization} (SPR) term
%
\begin{align*}
\footnotesize
&z(W;\alpha,M)=\\
&\begin{cases}
 2\sqrt{(1-\alpha)\alpha }||W||_2 & \text{if }   \frac{||W||_{\infty}}{M}\leq \sqrt{\frac{\alpha }{1-\alpha}}||W||_2\leq 1\\
\frac{\alpha M}{||W||_{\infty}}||W||^2_2+(1-\alpha)\frac{||W||_{\infty}}{M} &  \text{if }  \sqrt{\frac{\alpha }{1-\alpha}}||W||_2\leq   \frac{||W||_{\infty}}{M}\leq 1\\
 \alpha ||W||^2_2+(1-\alpha) & \text{otherwise,} 
\end{cases}
\end{align*}
where $M$ is a constant, $\alpha$ is a tunable hyper-parameter and $W$ is the weight tensor corresponding to the structure we want to prune (e.g., the weight matrix of a neuron). That is, in order to prune the ANN one trains it using as loss function
\[
 \textstyle
 L(X,Y,W) +\lambda\sum_{j\in \mathcal{N}}z(W_j;\alpha,M),
\]
where $W_j$ is the weight matrix corresponding to neuron $j$ and $\mathcal{N}$ is the set of neurons of the ANN. Coupled with a final magnitude-based pruning step, this approach has been shown to provide state-of-the-art pruning performances thanks to the unique and interesting properties of the SPR term. This potentially comes at the expense of extra hyperparameter tuning effort for $\alpha$ and $M$, which is unlikely to be a major issue in this application since ANNs that can be embedded in a MILP, even after pruning, cannot possibly have the extremely large size common in applications like Computer Vision and Naturale Language Processing, and therefore their training and tuning time is unlikely to be a major factor.

\section{Pruning as a Speed-Up Strategy}
\label{sec:pruningformip}
As previously mentioned, in the context of embedding ANNs in MIPs, scalability becomes a significant challenge as the number of (binary) variables and constraints grows proportionally with the number of parameters in the embedded ANN, but the cost of solving the MI(L)P may well grow exponentially in the number of (binary) variables. It therefore makes even more sense to employ the ML compression techniques that are used to reduce the computational resources required by ANNs. Many compression techniques other than pruning exist in the ML literature. However, not all of them are effective in the context of MIPs with embedded ANNs. For instance, quantization techniques aim to train networks that have weight values in a discrete (relatively small) set of $\mathbb{R}$ \cite{binconn,quant1}. One possibility is to directly implement the ANN using a lower bit number format than the standard Float-32 one \cite{myquant, quant2}. Quantization is a very popular technique in ML since can decrease both backward- and forward-pass computational effort, at the same time reducing the memory footprint of the resulting model. However, in the contest of MIPs, quantization does not bring any advantage, since the resulting problem from embedding a quantized ANN is not significantly different, from an Operations Research point of view, to the one where a non-quantized model has been embedded. Indeed, weights are coefficients in \eqref{scalprod}--\eqref{bin}, and having them in a small set of (integer) values may at most have a minor impact on the solution time. 
Other methods, like low-rank decomposition and parameter-sharing techniques \cite{lra1,lra2,eigen,fuj}, modify the internal operations of layers; this means that they cannot directly be used in this context without the development of new, specific formulations and new algorithms that can automatically detect them in a MIP problem. 

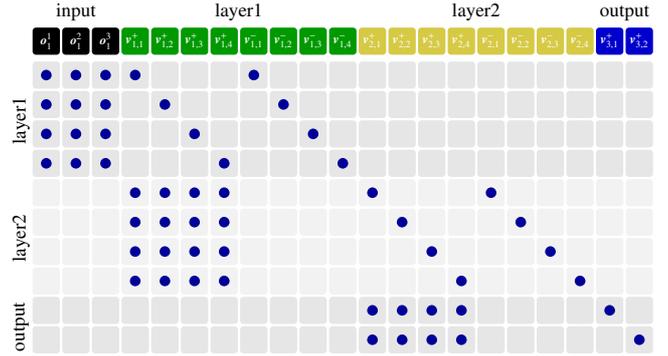
\begin{figure}
    \centering
\scalebox{0.6}{
\begin{tikzpicture}[mystyle/.style = {rounded corners=0.075cm},fontstyle/.style={white, font={\bfseries\fontsize{8}{0}\selectfont}}]
\def\L{21} 
\def\H{11} 
\def\dshift{0.6} 
\def\wspace{0.05} 

\foreach \i [evaluate={\strtx=(\i-1)*(\dshift+\wspace)};] in {1,...,\L}{
    \foreach \j [evaluate={\strty=-(\j-1)*(\dshift+\wspace)};] in {1,...,\H}{
    \ifnum\j=1
        \node (N\i_\j) at (\strtx+0.5*\dshift,\strty-0.5*\dshift) {};
    \else
        \node (N\i_\j) at (\strtx+0.5*\dshift,\strty-0.5*\dshift-2*\wspace) {};
    \fi
    }
}

\foreach \i in {1,...,\L}{
    \foreach \j in {2,...,5}{
    \fill[mygray, mystyle] ($(N\i_\j)+(-\dshift*0.5,\dshift*0.5)$) rectangle ($(N\i_\j)+(\dshift*0.5,-\dshift*0.5)$);
    }
}

\foreach \i in {1,...,\L}{
    \foreach \j in {6,...,9}{
    \fill[mygray!50!white, mystyle] ($(N\i_\j)+(-\dshift*0.5,\dshift*0.5)$) rectangle ($(N\i_\j)+(\dshift*0.5,-\dshift*0.5)$);
    }
}
\foreach \i in {1,...,\L}{
    \foreach \j in {10,11}{
    \fill[mygray, mystyle] ($(N\i_\j)+(-\dshift*0.5,\dshift*0.5)$) rectangle ($(N\i_\j)+(\dshift*0.5,-\dshift*0.5)$);
    }
}

\foreach \i in {1,...,\L}{
        \ifnum\i<4   
                \fill[black, mystyle] ($(N\i_1)+(-\dshift*0.5,\dshift*0.5)$) rectangle ($(N\i_1)+(\dshift*0.5,-\dshift*0.5)$);
        \else
            \ifnum\i<12
                \fill[mygreen, mystyle] ($(N\i_1)+(-\dshift*0.5,\dshift*0.5)$) rectangle ($(N\i_1)+(\dshift*0.5,-\dshift*0.5)$);
 
            \else
                \ifnum\i<20
                    \fill[myyellow, mystyle] ($(N\i_1)+(-\dshift*0.5,\dshift*0.5)$) rectangle ($(N\i_1)+(\dshift*0.5,-\dshift*0.5)$);

                \else
                    \fill[myblue, mystyle] ($(N\i_1)+(-\dshift*0.5,\dshift*0.5)$) rectangle ($(N\i_1)+(\dshift*0.5,-\dshift*0.5)$);

                \fi                
            \fi
        \fi
}

\foreach \i in {1,...,3}
    \node[fontstyle] at (N\i_1) {$\boldsymbol{o}_{1}^{\i}$};

\foreach[evaluate={\plus=int(\i+3);\minus=int(\i+7);}] \i in {1,...,4}{
    \node[fontstyle] at (N\plus_1) {$\boldsymbol{v}_{1,\i}^{+}$};
    \node[fontstyle] at (N\minus_1) {$\boldsymbol{v}_{1,\i}^{-}$};
}

\foreach[evaluate={\plus=int(\i+11);\minus=int(\i+15);}] \i in {1,...,4}{
    \node[fontstyle] at (N\plus_1) {$\boldsymbol{v}_{2,\i}^{+}$};
    \node[fontstyle] at (N\minus_1) {$\boldsymbol{v}_{2,\i}^{-}$};
}

\foreach[evaluate={\idx=int(\i+19)}] \i in {1,...,2}
    \node[fontstyle] at (N\idx_1) {$\boldsymbol{v}_{3,\i}^{+}$};

\foreach \i in {1,...,3}
    \foreach \j in {2,...,5}
        \node[blue!60!black, scale=5] at (N\i_\j) {$.$};
        
\foreach \i in {4,...,7}
    \foreach \j in {6,...,9}
        \node[blue!60!black, scale=5] at (N\i_\j) {$.$};

\foreach \i in {12,...,15}
    \foreach \j in {10,11}
        \node[blue!60!black, scale=5] at (N\i_\j) {$.$};

\foreach[evaluate={\idx=int(\i+2)}] \i in {2,...,5}
        \node[blue!60!black, scale=5] at (N\idx_\i) {$.$};
        
\foreach[evaluate={\idx=int(\i+6)}] \i in {2,...,9}
        \node[blue!60!black, scale=5] at (N\idx_\i) {$.$};
        
\foreach[evaluate={\idx=int(\i+10)}] \i in {6,...,11}
        \node[blue!60!black, scale=5] at (N\idx_\i) {$.$};

\node[above=0.3 cm] at (N2_1) {input};

\node[above=0.3 cm] at ($(N7_1)!0.5!(N8_1)$) {layer1};

\node[above=0.3 cm] at ($(N15_1)!0.5!(N16_1)$) {layer2};

\node[above=0.3 cm] at ($(N20_1)!0.5!(N21_1)$) {output};

\node[left=0.55 cm, rotate= 90] at ($(N1_2)!0.5!(N1_3)$) {layer1};

\node[left=0.55 cm, rotate= 90] at ($(N1_6)!0.5!(N1_7)$) {layer2};

\node[left=0.55 cm, rotate= 90] at ($(N1_9)!0.5!(N1_10)$) {output};

\end{tikzpicture}
}

    \caption{Constraints matrix}
    \label{fig:mat_normal}
\end{figure}

\begin{figure}
    \centering

\scalebox{0.6}{
\begin{tikzpicture}[x=2.7cm,y=1.6cm]
  \message{^^JNeural network activation}
  \def\NI{3} 
  \def\NH{4} 
  \def\NN{4} 
  \def\NO{2} 
  \def\yshift{0.0} 
  
  \foreach \i [evaluate={\c=int(\i==\NI); \y=\NI/2-\i-\c*\yshift; \index=(\i<\NI?int(\i):"n");}]
              in {1,...,\NI}{ 
    \node[node black,outer sep=0.6] (NI-\i) at (0,\y) {};
  }

  \foreach \i [evaluate={\c=int(\i==\NH); \y=\NH/2-\i-\c*\yshift; \index=(\i<\NH?int(\i):"m");}]
    in {\NH,...,1}{ 
      \node[node green]
        (NH-\i) at (1,\y) {};
      \foreach \j in {1,...,\NI}{ 
        \draw[connect,myblue!20] (NI-\j) -- (NH-\i);
      }
  }

  \foreach \i [evaluate={\c=int(\i==\NN); \y=\NN/2-\i-\c*\yshift; \index=(\i<\NN?int(\i):"m");}]
    in {\NN,...,1}{ 
    \ifnum\i=-1 
      \node[node yellow]
        (NN-\i) at (2,\y) {};
      \foreach \j [evaluate={\index=(\j<\NH?int(\j):"n");}] in {1,...,\NH}{ 
        \draw[connect,white,line width=1.2] (NH-\j) -- (NN-\i);
        \draw[connect] (NH-\j) -- (NN-\i)
          node[pos=0.50] {\contour{white}{}};
      }
    \else 
      \node[node yellow]
        (NN-\i) at (2,\y) {};
      \foreach \j in {1,...,\NH}{ 
        \draw[connect,myblue!20] (NH-\j) -- (NN-\i);
      }
    \fi
  }

  \foreach \i [evaluate={\c=int(\i==\NO); \y=\NO/2-\i-\c*\yshift; \index=(\i<\NO?int(\i):"m");}]
    in {\NO,...,1}{ 
   
      \node[node blue]
        (NO-\i) at (3,\y) {};
      \foreach \j [evaluate={\index=(\j<\NN?int(\j):"n");}] in {1,...,\NN}{ 
      \ifnum\j=-1 
        \draw[connect,white,line width=1.2] (NN-\j) -- (NO-\i);
        \draw[connect] (NN-\j) -- (NO-\i)
          node[pos=0.50] {\contour{white}{}};

    \else 
        \draw[connect,myblue!20] (NN-\j) -- (NO-\i);
    \fi
    }
  }
\end{tikzpicture}
}

    \caption{Corresponding network}
    \label{fig:net_unpruned}
\end{figure}
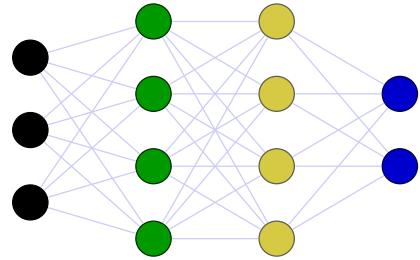

By contrast, structured pruning techniques perfectly fit the needs of embedding an ANN in a MIP. Even unstructured pruning may have some impact, since when a weight is removed (i.e., set to zero) the corresponding entry in the MIP constraints matrix is also set to zero, leading to a sparser constraints matrix. However, entirely removing variables or constraints is more effective; in the case of a feed-forward ANN, this corresponds to performing structured pruning on neurons, as visualized in Figures~\ref{fig:mat_normal}-~\ref{fig:net_unpruned}-~\ref{fig:mat_pruned}-~\ref{fig:net_pruned}.


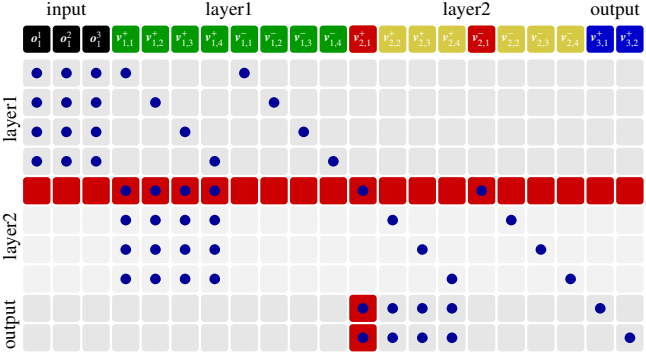
\begin{figure}
    \centering

\scalebox{0.6}{
\begin{tikzpicture}[mystyle/.style = {rounded corners=0.075cm},fontstyle/.style={white, font={\bfseries\fontsize{8}{0}\selectfont}}]

\def\L{21} 
\def\H{11} 
\def\dshift{0.6} 
\def\wspace{0.05} 

\foreach \i [evaluate={\strtx=(\i-1)*(\dshift+\wspace)};] in {1,...,\L}{
    \foreach \j [evaluate={\strty=-(\j-1)*(\dshift+\wspace)};] in {1,...,\H}{
    \ifnum\j=1
        \node (N\i_\j) at (\strtx+0.5*\dshift,\strty-0.5*\dshift) {};
    \else
        \node (N\i_\j) at (\strtx+0.5*\dshift,\strty-0.5*\dshift-2*\wspace) {};
    \fi
    }
}

\foreach \i in {1,...,\L}{
    \foreach \j in {2,...,5}{
    \fill[mygray, mystyle] ($(N\i_\j)+(-\dshift*0.5,\dshift*0.5)$) rectangle ($(N\i_\j)+(\dshift*0.5,-\dshift*0.5)$);
    }
}

\foreach \i in {1,...,\L}{
    \foreach \j in {6,...,9}{
    \fill[mygray!50!white, mystyle] ($(N\i_\j)+(-\dshift*0.5,\dshift*0.5)$) rectangle ($(N\i_\j)+(\dshift*0.5,-\dshift*0.5)$);
    }
}
\foreach \i in {1,...,\L}{
    \foreach \j in {10,11}{
    \fill[mygray, mystyle] ($(N\i_\j)+(-\dshift*0.5,\dshift*0.5)$) rectangle ($(N\i_\j)+(\dshift*0.5,-\dshift*0.5)$);
    }
}

\foreach \i in {1,...,\L}{
    \fill[myred, mystyle] ($(N\i_6)+(-\dshift*0.5,\dshift*0.5)$) rectangle ($(N\i_6)+(\dshift*0.5,-\dshift*0.5)$);
    }

\fill[myred, mystyle] ($(N12_1)+(-\dshift*0.5,\dshift*0.5)$) rectangle ($(N12_1)+(\dshift*0.5,-\dshift*0.5)$);

\foreach \i in {1,...,\L}{
        \ifnum\i<4   
                \fill[black, mystyle] ($(N\i_1)+(-\dshift*0.5,\dshift*0.5)$) rectangle ($(N\i_1)+(\dshift*0.5,-\dshift*0.5)$);
        \else
            \ifnum\i<12
                \fill[mygreen, mystyle] ($(N\i_1)+(-\dshift*0.5,\dshift*0.5)$) rectangle ($(N\i_1)+(\dshift*0.5,-\dshift*0.5)$);
 
            \else
                \ifnum\i<20
                    \fill[myyellow, mystyle] ($(N\i_1)+(-\dshift*0.5,\dshift*0.5)$) rectangle ($(N\i_1)+(\dshift*0.5,-\dshift*0.5)$);

                \else
                    \fill[myblue, mystyle] ($(N\i_1)+(-\dshift*0.5,\dshift*0.5)$) rectangle ($(N\i_1)+(\dshift*0.5,-\dshift*0.5)$);

                \fi                
            \fi
        \fi
}
\fill[myred, mystyle] ($(N12_1)+(-\dshift*0.5,\dshift*0.5)$) rectangle ($(N12_1)+(\dshift*0.5,-\dshift*0.5)$);

\fill[myred, mystyle] ($(N16_1)+(-\dshift*0.5,\dshift*0.5)$) rectangle ($(N16_1)+(\dshift*0.5,-\dshift*0.5)$);

\fill[myred, mystyle] ($(N12_10)+(-\dshift*0.5,\dshift*0.5)$) rectangle ($(N12_10)+(\dshift*0.5,-\dshift*0.5)$);
\fill[myred, mystyle] ($(N12_11)+(-\dshift*0.5,\dshift*0.5)$) rectangle ($(N12_11)+(\dshift*0.5,-\dshift*0.5)$);
\foreach \i in {1,...,3}
    \node[fontstyle] at (N\i_1) {$\boldsymbol{o}_{1}^{\i}$};

\foreach[evaluate={\plus=int(\i+3);\minus=int(\i+7);}] \i in {1,...,4}{
    \node[fontstyle] at (N\plus_1) {$\boldsymbol{v}_{1,\i}^{+}$};
    \node[fontstyle] at (N\minus_1) {$\boldsymbol{v}_{1,\i}^{-}$};
}

\foreach[evaluate={\plus=int(\i+11);\minus=int(\i+15);}] \i in {1,...,4}{
    \node[fontstyle] at (N\plus_1) {$\boldsymbol{v}_{2,\i}^{+}$};
    \node[fontstyle] at (N\minus_1) {$\boldsymbol{v}_{2,\i}^{-}$};
}

\foreach[evaluate={\idx=int(\i+19)}] \i in {1,...,2}
    \node[fontstyle] at (N\idx_1) {$\boldsymbol{v}_{3,\i}^{+}$};

\foreach \i in {1,...,3}
    \foreach \j in {2,...,5}
        \node[blue!60!black, scale=5] at (N\i_\j) {$.$};
        
\foreach \i in {4,...,7}
    \foreach \j in {6,...,9}
        \node[blue!60!black, scale=5] at (N\i_\j) {$.$};

\foreach \i in {12,...,15}
    \foreach \j in {10,11}
        \node[blue!60!black, scale=5] at (N\i_\j) {$.$};

\foreach[evaluate={\idx=int(\i+2)}] \i in {2,...,5}
        \node[blue!60!black, scale=5] at (N\idx_\i) {$.$};
        
\foreach[evaluate={\idx=int(\i+6)}] \i in {2,...,9}
        \node[blue!60!black, scale=5] at (N\idx_\i) {$.$};
        
\foreach[evaluate={\idx=int(\i+10)}] \i in {6,...,11}
        \node[blue!60!black, scale=5] at (N\idx_\i) {$.$};

\node[above=0.3 cm] at (N2_1) {input};

\node[above=0.3 cm] at ($(N7_1)!0.5!(N8_1)$) {layer1};

\node[above=0.3 cm] at ($(N15_1)!0.5!(N16_1)$) {layer2};

\node[above=0.3 cm] at ($(N20_1)!0.5!(N21_1)$) {output};

\node[left=0.55 cm, rotate= 90] at ($(N1_2)!0.5!(N1_3)$) {layer1};

\node[left=0.55 cm, rotate= 90] at ($(N1_6)!0.5!(N1_7)$) {layer2};

\node[left=0.55 cm, rotate= 90] at ($(N1_9)!0.5!(N1_10)$) {output};

\end{tikzpicture}

  }

    \caption{Constraints matrix, in red the removed constraints and variables due to neurons pruning}
    \label{fig:mat_pruned}
\end{figure}
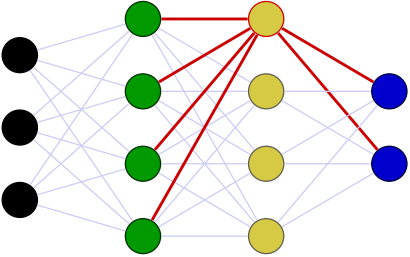
\begin{figure}
    \centering

\scalebox{0.6}{
\begin{tikzpicture}[x=2.7cm,y=1.6cm]
  \message{^^JNeural network activation}
  \def\NI{3} 
  \def\NH{4} 
  \def\NN{4} 
  \def\NO{2} 
  \def\yshift{0.0} 
  
  \foreach \i [evaluate={\c=int(\i==\NI); \y=\NI/2-\i-\c*\yshift; \index=(\i<\NI?int(\i):"n");}]
              in {1,...,\NI}{ 
    \node[node black,outer sep=0.6] (NI-\i) at (0,\y) {};
  }

  \foreach \i [evaluate={\c=int(\i==\NH); \y=\NH/2-\i-\c*\yshift; \index=(\i<\NH?int(\i):"m");}]
    in {\NH,...,1}{ 
      \node[node green]
        (NH-\i) at (1,\y) {};
      \foreach \j in {1,...,\NI}{ 
        \draw[connect,myblue!20] (NI-\j) -- (NH-\i);
      }
  }

  \foreach \i [evaluate={\c=int(\i==\NN); \y=\NN/2-\i-\c*\yshift; \index=(\i<\NN?int(\i):"m");}]
    in {\NN,...,1}{ 
    \ifnum\i=1 
      \node[node yellow,draw=myred]
        (NN-\i) at (2,\y) {};
      \foreach \j [evaluate={\index=(\j<\NH?int(\j):"n");}] in {1,...,\NH}{ 
        \draw[connect,myred,line width=1.8] (NH-\j) -- (NN-\i);
        \draw[connect,myred,line width=0.05] (NH-\j) -- (NN-\i)
          node[pos=0.50] {\contour{myred}{}};
      }
    \else 
      \node[node yellow]
        (NN-\i) at (2,\y) {};
      \foreach \j in {1,...,\NH}{ 
        \draw[connect,myblue!20] (NH-\j) -- (NN-\i);
      }
    \fi
  }

  \foreach \i [evaluate={\c=int(\i==\NO); \y=\NO/2-\i-\c*\yshift; \index=(\i<\NO?int(\i):"m");}]
    in {\NO,...,1}{ 
   
      \node[node blue]
        (NO-\i) at (3,\y) {};
      \foreach \j [evaluate={\index=(\j<\NN?int(\j):"n");}] in {1,...,\NN}{ 
      \ifnum\j=1 
        \draw[connect,myred,line width=1.8] (NN-\j) -- (NO-\i);
        \draw[connect,myred,line width=0.05] (NN-\j) -- (NO-\i)
          node[pos=0.50] {\contour{myred}{}};

    \else 
        \draw[connect,myblue!20] (NN-\j) -- (NO-\i);
    \fi
    }
  }
\end{tikzpicture}
}

    \caption{Corresponding network, the highlighted part is the pruned one.}
    \label{fig:net_pruned}
\end{figure}

It is interesting to remark that, for ML purposes, pruning techniques only bring advantages at inference time, but reducing the number of parameters only reduces linearly the computational cost of the forward pass. By contrast, removing neurons of a network brings an exponential speed up in the time required to solve the resulting MIP formulations. Hence, pruning is arguably more relevant for OR than for ML, despite having been developed in the latter area. In particular, structured pruning---as opposed to unstructured one---is crucial in that it allows using existing automatic structure detection algorithms, such as that implemented in Gurobi, while unstructured pruning is very likely to result in a different structure of the constraints matrix that would not be recognizable, thereby preventing the use of OBBT techniques that are crucial in this context.

Based on the considerations above, we argue about modifying the existing pipeline for embedding ANNs in MIPs. After training the ANN (or during training, depending on the technique used), we prune the model before embedding it in the MIP formulation of the problem in hand. This approach either reduces the solution time of the MIP with the same generalisation performances, or, possibly, allows one to include larger,  therefore more expressive ANNs, capable of achieving higher accuracy while still maintaining the ability to solve the resulting MIPs within a reasonable time. In particular, we will employ the Structured Perspective Regularization, i.e., we train the ANN by adding the SPR term to the loss, which will lead to a weight tensor with a structured sparsity. After fixing to zero (i.e., removing) neurons whose weights are all below a fixed threshold, we fine-tune the network with a standard loss for a few more epochs (see ~\cite{spr} for details). The obtained ANN is then embedded in the MIP, and it will require the addition of fewer variables and constraints with respect to its unpruned counterpart. 

\section{Experiments}
\label{sec:experiments}
\subsection{Building adversarial examples}

We test the effectiveness of pruning in the task of finding an adversarial example of a given network. In particular, we focus on the verification problem \cite{NetVer}, which consists in finding a \emph{slight} modification of an input that is originally correctly classified by the network in such a way that the modified one is assigned to a chosen class by the ANN. More formally, assume we are given a trained ANN $\bar{g}(\cdot):\mathbb{R}^n\rightarrow [0,1]^C$ and one input $x$ such that $\bar{g}(x)$ has its maximum value at the coordinate corresponding to the correct class of $x$. Denoting with $k$ this coordinate and with $h$ the coordinate with the second highest value of $\bar{g}(x)$, the problem we want to solve is
\begin{align}
    \max\;& y_h-y_k \label{adv1}\\
    \text{s.t.}\;
    & y =\bar{g}(\bar{x}) \label{nn_constr} \\
    & \Delta \geq x-\bar{x} \\
    & \Delta \geq \bar{x}-x \\
    & \bar{x} \in \mathbb{R}^n \label{adv2}
\end{align}
where $\Delta$ is a given distance bound. Clearly, \eqref{adv1}--\eqref{adv2} is a special case of the MPLC class \eqref{eq:genof}--\eqref{eq:int}. In particular, the constraint \eqref{nn_constr} encodes an ANN function, so it needs to be handled with the techniques we presented in Section~\ref{section:embed}. We selected this problem since it is of great interest to ML researchers. Furthermore, it can in principle be relevant to test the robustness of networks of any size, and therefore it allows to explore the boundaries of what MPLC approaches (with or without pruning) can achieve.

\subsection{General setup and notation}

To test the effectiveness of our pruning techniques, we ran some experiments on network robustness using the MNIST dataset. We used the same settings of the notebook available at \url{https://github.com/Gurobi/gurobi-machinelearning/blob/main/notebooks/adversarial/adversarial_pytorch.ipynb}, where formulation \eqref{adv1}-\eqref{adv2} is solved with $\Delta=5$.

We trained the ANNs using the Pytorch SGD optimizer with no weight decay and no momentum. We used 128 as batch size and we trained the network for 50 epochs with a constant learning rate equal to $0.1$. All the networks are Pytorch sequential models containing only Linear and ReLU layers. For the pruned networks, we performed a (limited) 3 by 3 grid search to choose the $\lambda$ factor that multiplies the SPR term and the $\alpha$ hyper-parameter needed in its definition ($M$ is automatically set as in \cite{spr}). After 50 training epochs, the model is fine-tuned for 10 epochs without using any regularization. Note that the objective of the grid search is to find the smallest network that keeps basically the same out-of-sample accuracy of the original one, and better results could conceivably be obtained by employing end-to-end techniques that take into account the optimization process in the computation of the loss \cite{spo,ltr}.

In tables \ref{tab:delta5} and \ref{tab:delta20}, the first column reports the network architecture of the used ANN and if pruning was used, while the $\Delta$ parameter value of \eqref{adv1}-\eqref{adv2} can be found in the first row. We compare the result of the baseline approach (i.e., without pruning) and the result obtained using the pruning method with the best hyper-parameters found. We report the validation accuracy (in percentage), the time needed by Gurobi to solve the obtained MIP (in seconds), and the number of branch-and-bound nodes explored during that time. Additionally, for the pruned networks, we report the value of $\lambda$ and $\alpha$ and the architecture of the network after pruning. When referring to a network architecture, the terms $L$x$N$ refer to a sequence of $L$ layers each of them containing $N$ neurons. When multiple terms follow each other, it indicates their order in the network. For example, 2x20-3x10 stands for a network that starts with 2 layers of 20 neurons and continues with 3 layers of 10 neurons. Each experiment is repeated 3 times and a time limit of 1800 seconds is given to Gurobi.

\subsection{Detailed results}

 Table~\ref{tab:delta5} shows the results using $\Delta$=5 on 4 different architectures with an increasing number of neurons and layers. When pruning small architectures, like the 2x50 and 2x100 networks, pruning the ANN results in at least halving the time used by Gurobi. Moreover, the accuracy of the pruned models is higher than the baseline, this is, likely, because pruning has also a regularization effect.

The results on the 2x200 architecture show that the baseline is not able to solve the problems in the given time for two out of three runs. Instead, our method always leads to MIPs that are easily solved by Gurobi while maintaining the same accuracy as the baseline.

Finally, we report the results using the 6x100 networks, significantly bigger with respect to the previous ones. The baseline, once again, cannot solve two out of the three problems in the given time limit. Instead, our method is able to succeed in all cases, at the cost of losing a little bit of accuracy (0.3 percent in the best case).

As a last remark, we notice that for all the MIPs we solved relatively to unpruned network, no counterexample existed in the given neighborhood (i.e., the optimal value of \eqref{adv1}-\eqref{adv2} is negative). This remains true for the corresponding pruned counterparts, confirming that the pruned and unpruned versions of the MIPs are qualitatively very similar.

\begin{table}[!ht]
    \centering
    \scalebox{0.7}{
    \begin{tabular}{rrrrrr}
            \toprule
        \multicolumn{6}{c}{$\Delta=5$}\\
    \toprule
        Arch & $\lambda$-$\alpha$ & Acc. & Time & Nodes & Pruned Arch \\ 
        \midrule
        \multirow{3}{*}{\begin{tabular}{@{}c@{}}2x50\\Baseline\end{tabular}}  & ~ & 97.55 & 14.88 & 5820 & ~ \\ 
        ~ & ~ & 97.47 & 20.65 & 12040 & ~ \\ 
        ~ & ~ & 97.25 & 8.07 & 9497 & ~ \\
        \midrule
        \multirow{3}{*}{\begin{tabular}{@{}c@{}}2x50\\Pruned\end{tabular}} & \multirow{3}{*}{0.5-0.9} & 97.77 & 3.29 & 3328 & 1x39-1x43 \\ 
           & ~ & 97.49 & 3.93 & 6482 & 1x30-1x42 \\ 
         & ~ & 97.73 & 1.96 & 3992 & 1x39-1x42\\ 
    \toprule
        \multirow{3}{*}{\begin{tabular}{@{}c@{}}2x100\\Baseline\end{tabular}} & ~ & 97.96 & 39.29 & 2971 &   \\ 
         ~ & ~ & 97.76 & 35.01 & 3112 &   \\ 
         ~ & ~ & 97.97 & 39.00 & 3019 &   \\ \midrule
        \multirow{3}{*}{\begin{tabular}{@{}c@{}}2x100\\Pruned\end{tabular}} & \multirow{3}{*}{0.5-0.9} & 98.08 & 15.99 & 3066 & 1x61-1x80 \\ 
         ~ & ~ & 98.01 & 15.65 & 3107 & 1x61-1x87 \\ 
        ~ & ~ & 98.04 & 17.67 & 2951 & 1x63-1x86 \\  
    \toprule
        \multirow{3}{*}{\begin{tabular}{@{}c@{}}2x200\\Baseline\end{tabular}} &  ~ & 98.14 & 1800.37 & 424758 &   \\ 
        &  ~ & 98.04 & 1800.18 & 401361 &   \\ 
        ~ &  ~ & 97.95 & 781.90 & 58656 &    \\ \midrule
        \multirow{3}{*}{\begin{tabular}{@{}c@{}}2x200\\Pruned\end{tabular}} & \multirow{3}{*}{0.5-0.5} & 97.96 & 18.66 & 3029 & 1x56-1x144 \\ 
        ~ &  ~ & 98.13 & 24.65 & 3600 & 1x57-1x144 \\ 
        ~ &  ~ & 98.04 & 28.19 & 2997 & 1x59-1x140 \\ 
    \toprule
        \multirow{3}{*}{\begin{tabular}{@{}c@{}}6x100\\Baseline\end{tabular}} & ~ & 97.60 & 474,76 & 15261 &   \\ 
        ~ & ~ &97.77 & 1800.02 & 798306 &  \\ 
        ~ &  ~ & 97.67 & 818.19 & 14334 &  \\ \midrule
        \multirow{6}{*}{\begin{tabular}{@{}c@{}}6x100\\Pruned\end{tabular}} & \multirow{6}{*}{1.0-0.1} & \multirow{2}{*}{97.52} & \multirow{2}{*}{231.02} & \multirow{2}{*}{3173} & 1x39-1x82 \\ 
         & ~ &  ~ & ~ & ~ & 2x61-1x60-1x54  \\ 
        ~ & ~ & \multirow{2}{*}{97.47} & \multirow{2}{*}{79.22} & \multirow{2}{*}{7566} & 1x44-1x71-1x49\\ 
         & ~ &  ~ & ~ & ~ & -1x53-1x49-1x45 \\ 
        ~ &  ~ & \multirow{2}{*}{97.21} & \multirow{2}{*}{44.24} & \multirow{2}{*}{11417} & 1x37-1x72-1x48 \\ 
         & ~ & ~ & ~ & ~ & -1x51-1x48-1x46 \\ 
        \bottomrule
    \end{tabular}
    }
            \caption{Results using $\Delta=5$.
            }
            \label{tab:delta5}
\end{table}

\subsection{Investigating the quality of the solutions}

To better validate the quality of our results, we solved again the adversarial problem \eqref{adv1}-\eqref{adv2} using $\Delta$=20 and employing the same networks trained in the previous experiments. This was aimed to find adversarial examples in the given region to better understand the effect of pruning on the resulting MIP. We report the results in Table~\ref{tab:delta20}, where the  ``accuracy" and  ``pruned architecture" columns have been removed since they are the same as in the previous table. For all the experiments, a counter-example existed in the given region, and in the last column of Table~\ref{tab:delta20}, named  ``Found", we report if Gurobi was able to find one adversarial example in the given time limit. Unsurprisingly, for all the MIPs corresponding to pruned networks, Gurobi was able to find an adversarial example within a time considerably inferior to the 1800 seconds limit. Moreover, all the adversarial examples obtained using a pruned network were also adversarial for the unpruned counterpart with the same starting architecture. This empirically proved that, in our setting, pruning can be even used to solve the adversarial example problem for the unpruned counterpart and it is again a good indication that pruning does not heavily affect the resulting MIP. This is in accordance with the ML literature, where there is a good consensus that not-too-aggressive pruning of ANNs does not significantly impacts their robustness \cite{robustpruning,robustpruning2}, and therefore the existence---or not---of the counter-example in our application. Finally, the times reported in Table~\ref{tab:delta20} show that the speed-up is still very significant even with the new value of $\Delta$ and that in some cases the Baseline is not able to find any adversarial example.

We conclude this section by noting that additional experiments, which are not included in this paper for the sake of brevity, have shown that a high setting of the OBBT parameter \cite{fischettiJo} of Gurobi is crucial to obtain good performances both for pruned and unpruned instances, confirming the importance of structured pruning.

\begin{table}[!ht]
    \centering
    \scalebox{0.7}{
    \begin{tabular}{rrrrr}
                \toprule
        \multicolumn{5}{c}{$\Delta=20$}\\
    \toprule
        Arch & $\lambda$-$\alpha$ & Time & Nodes & Found\\ \midrule
        \multirow{3}{*}{\begin{tabular}{@{}c@{}}2x50\\Baseline\end{tabular}} & ~  & 1.85  &  1 & YES\\ 
        ~ & ~  & 5.06 & 1221 & YES \\ 
        ~ &  ~ & 1.66 & 1 & YES \\
        \midrule
        \multirow{3}{*}{\begin{tabular}{@{}c@{}}2x50\\Pruned\end{tabular}} & \multirow{3}{*}{0.5-0.9}  & 2.73  & 127& YES  \\ 
          & ~ & 2.90 &  1128 & YES \\ 
         & ~  & 0.64 & 1 & YES \\
    \toprule
        \multirow{3}{*}{\begin{tabular}{@{}c@{}}2x100\\Baseline\end{tabular}}  & ~  & 17.35  & 1217 & YES  \\ 
         ~ & ~  & 147.67 & 7532 & YES \\ 
         ~ & ~ & 102.67 &  3252 & YES\\
        \midrule
        \multirow{3}{*}{\begin{tabular}{@{}c@{}}2x100\\Pruned\end{tabular}} & \multirow{3}{*}{0.5-0.9}  & 6.66  & 2079  & YES\\ 
           & ~ & 6.03 &  127 & YES\\ 
          & ~  & 2.16 & 1 & YES\\ 
    \toprule
        \multirow{3}{*}{\begin{tabular}{@{}c@{}}2x200\\Baseline\end{tabular}}  & ~  & 439.17  &  40075 & YES\\ 
         ~ & ~  & 563.24 & 17597 & YES\\ 
         ~ & ~ & 508.14 & 6014 & YES\\
        \midrule
        \multirow{3}{*}{\begin{tabular}{@{}c@{}}2x200\\Pruned\end{tabular}} & \multirow{3}{*}{0.5-0.5}  & 2.56 & 1  & YES\\ 
           & ~ & 18.65 & 5433  & YES\\ 
         & ~  & 9.60 & 1202 & YES\\
    \toprule
        \multirow{3}{*}{\begin{tabular}{@{}c@{}}6x100\\Baseline\end{tabular}}  & ~  & 1800.06  &  138918 & NO\\ 
         ~ & ~  & 1800.03 & 237328 & NO\\ 
        ~ &  ~ & 1800.10 & 184954 & NO\\
        \midrule
        \multirow{3}{*}{\begin{tabular}{@{}c@{}}6x100\\Pruned\end{tabular}} & \multirow{3}{*}{1.0-0.1}  & 15.53 &  1 & YES\\ 
           & ~ & 7.51 &  1 & YES\\ 
          & ~  & 129.70 & 27045 & YES\\ \bottomrule
    \end{tabular}
    }
        \caption{Results using $\Delta=20$. 
        }
        \label{tab:delta20}
\end{table}

\section{Conclusions and future directions}

This paper has demonstrated the effectiveness of pruning artificial neural networks in accelerating the solution time of mixed-integer programming problems that incorporate ANNs. The choice of the sparsity structure for pruning plays a crucial role in achieving significant speed-up, and we argued that structured pruning is superior to unstructured one. 
Further research in this area can focus on gaining a deeper understanding of which sparsity structures are most suitable for improving the solution time of MIPs. Exploring the trade-off between pruning-induced sparsity and solution quality is another interesting avenue for future investigations. By advancing our understanding of pruning techniques and their impact on MIPs, we can enhance the efficiency and scalability of embedding ANNs in optimization problems.

\section*{Acknowledgments}
The authors are grateful to Pierre Bonami for his generous and insightful feedback.
This work has been supported by the NSERC Alliance grant 544900- 19 in collaboration with Huawei-Canada

\bibliographystyle{elsarticle-num} 
\bibliography{mybib.bib}

\end{document}